\documentclass[11pt]{article}

\usepackage[preprint]{acl}

\usepackage{array}
\usepackage{booktabs}
\usepackage{adjustbox}
\usepackage{multirow}
\usepackage{amsmath}
\usepackage{amssymb}

\usepackage{color}
\usepackage{tcolorbox}          
\tcbuselibrary{skins,breakable} 
\usepackage[table]{xcolor}

\definecolor{tred}{RGB}{251, 130, 132}
\newcommand{\highlightbox}[3]{
\begin{tcolorbox}[
    enhanced jigsaw,
    breakable,
    pad at break*=1mm,
    colback=#1!5!white,
    colframe=#1,
    title=#2
]
#3
\end{tcolorbox}
}

\newcommand{\tb}[1]{\textbf{#1}}

\usepackage{times}
\usepackage{latexsym}

\usepackage[T1]{fontenc}

\usepackage[utf8]{inputenc}

\usepackage{microtype}

\usepackage{inconsolata}

\usepackage{graphicx}
\usepackage{enumitem} 

\newcommand{\yh}[1]{\textcolor{black}{#1}}

\newcommand{\zzy}[1]{\textcolor{black}{#1}}

\title{TingIS: Real-time Risk Event Discovery from Noisy Customer Incidents\\
at Enterprise Scale}

\author{
    Jun Wang\textsuperscript{1}\thanks{Equal contribution.} \quad
    Ziyin Zhang\textsuperscript{1,2$*$} \quad
    Rui Wang\textsuperscript{1} \quad
    Hang Yu\textsuperscript{1\textdagger} \quad
    Peng Di\textsuperscript{1\textdagger} \quad
    Rui Wang\textsuperscript{2\textdagger} \\
    \textsuperscript{1}Ant Group \\
    \textsuperscript{2}Shanghai Jiao Tong University \\
    \normalsize{\textsuperscript{\textdagger}\texttt{hyu1@e.ntu.edu.sg, dipeng.dp@antgroup.com, wangrui12@sjtu.edu.cn}}
}

\AtBeginDocument{
\setlength{\abovedisplayskip}{0.6ex}
\setlength{\belowdisplayskip}{0.6ex}
}
\allowdisplaybreaks[4]

\begin{document}
\maketitle
\begin{abstract}
Real-time detection and mitigation of technical anomalies are critical for large-scale cloud-native services, where even minutes of downtime can result in massive financial losses and diminished user trust. While customer incidents serve as a vital signal for discovering risks missed by monitoring, extracting actionable intelligence from this data remains challenging due to extreme noise, high throughput, and semantic complexity of diverse business lines. In this paper, we present TingIS, an end-to-end system designed for enterprise-grade incident discovery. At the core of TingIS is a multi-stage event linking engine that synergizes efficient indexing techniques with Large Language Models (LLMs) to make informed decisions on event merging, enabling the stable extraction of actionable incidents from just a handful of diverse user descriptions. This engine is complemented by a cascaded routing mechanism for precise business attribution and a multi-dimensional noise reduction pipeline that integrates domain knowledge, statistical patterns, and behavioral filtering. Deployed in a production environment handling a peak throughput of over 2,000 messages per minute and 300,000 messages per day, TingIS achieves a P90 alert latency of 3.5 minutes and a 95\% discovery rate for high-priority incidents. Benchmarks constructed from real-world data demonstrate that TingIS significantly outperforms baseline methods in routing accuracy, clustering quality, and Signal-to-Noise Ratio\footnote{Code is released at \url{https://github.com/Geralt-Targaryen/CIIS}.}.
\end{abstract}

\section{Introduction}\label{sec:introduction}

In the era of modern digital services, large-scale online platforms - underpinned by complex microservices and cloud-native architectures - have become indispensable, powering everything from global e-commerce and social media to financial transactions. For these systems, even minor failures can rapidly propagate into large-scale incidents, causing significant financial losses and erosion of user trust. For instance, Alipay - one of the world’s largest mobile payment platforms - experienced a critical configuration error related to China's national subsidies in January 2025, where a 20\% discount is mistakenly applied to all transactions~\citep{alipay-error}. With an annual transaction volume of approximately \$20 trillion, even a 5-minute window for such an incident could result in an estimated loss of 40 million dollars~\citep{alipay-stats}. Thus, timely detection and response to such emerging risks are critical for maintaining system reliability and financial safety in practice. 


While internal observability systems such as metrics, logs, and traces form the first line of defense, they are not infallible. When they do fail, customer incidents such as online feedback and hotline inquiries provide a complementary and uniquely valuable signal, exposing failures in the ``blind spots'' of automated monitoring and reflecting a direct measure of user-perceived impact. Therefore, the early detection of latent system vulnerabilities - which we call ``\emph{risk events}'' - from as few as 3 customer incidents has emerged as a cornerstone strategy for preempting catastrophic failures and minimizing enterprise losses. However, leveraging customer incidents for real-time risk detection presents formidable challenges, as they are \tb{noisy, colloquial, and multi-source} by nature. \yh{Extracting a systemic failure signal from just 3 noisy data points amidst a streaming throughput of 2,000 messages per minute creates a severe Signal-to-Noise Ratio (SNR) challenge. A system with a low SNR would inevitably trigger thousands of false positive alerts, rapidly overwhelming Site Reliability Engineering (SRE) teams and leading to alert fatigue.} The situation is further complicated by \tb{business heterogeneity, high demand for real-timeness, and low tolerance to undetected failures}. 

\begin{figure*}[th]
    \centering
    \includegraphics[width=0.9\textwidth]{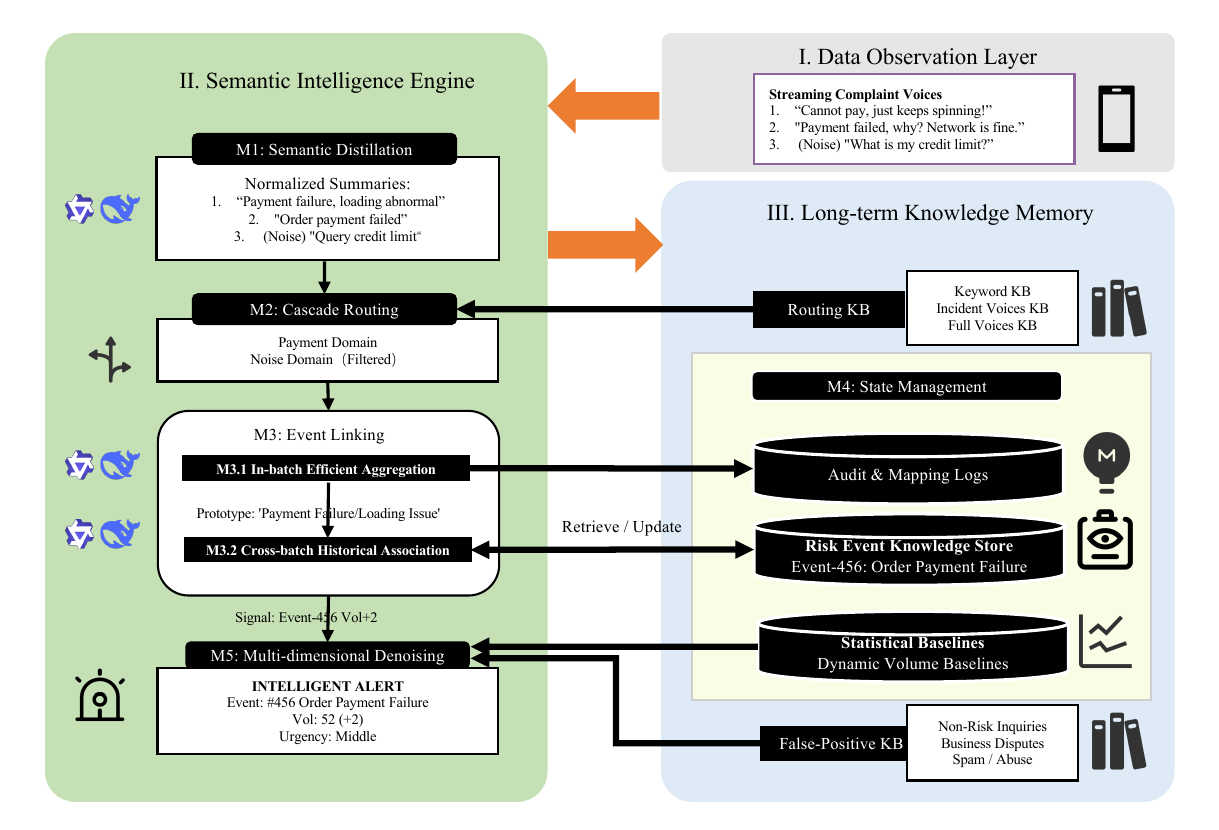}
    \caption{System architecture of TingIS, consisting of five modules (semantic distillation, cascaded routing, event linking, state management, and multi-dimensional denoising) across three layers (data observation, semantic engine, and long-term memory).}
    \label{fig:architecture}
\end{figure*}

In response to these challenges, we present TingIS (Ting Intelligent Service), an end-to-end system for mining risk events from customer incidents in large-scale production environments. Central to TingIS is a multi-stage event linking engine, which serves as the primary intelligence layer for synthesizing fragmented customer incidents into structured risk events. By synergizing Locality-Sensitive Hashing (LSH), historical event association, and the advanced reasoning of LLMs, this engine effectively bridges the gap between raw, noisy semantic inputs and actionable risk intelligence. This core capability is supported by four auxiliary modules - semantic distillation, cascaded routing, event state management, and multi-dimensional denoising - which together ensure the system maintains high accuracy, low latency, robust throughput, and low-effort maintainability in complex enterprise settings.

TingIS has been deployed on a leading financial technology platform, processing over 300,000 customer incidents daily with a peak throughput exceeding 2,000 incidents per minute. During a one-month online deployment, the system successfully identified 95\% of high-priority risk incidents with a P90 alert latency of 3.5 minutes, providing a critical window for rapid emergency response. Furthermore, extensive evaluations on benchmarks constructed from real-world production data demonstrate that TingIS significantly outperforms both system-level baselines and specialized module-level methods in terms of routing accuracy, clustering quality, and signal-to-noise ratio.

\section{System Architecture}

\highlightbox{tred}{Customer Incident}{
A \tb{customer incident} is an atomic unit of external feedback (e.g., a user complaint log). It is characterized as noisy, colloquial, and subjective.
}
\highlightbox{tred}{Risk Event}{
A \tb{risk event} is a structured representation of a system vulnerability or failure, \zzy{uniquely identified by a tuple of business domain attribution (\emph{biz\_code}) and topic (verified by SREs)}. Unlike an incident, a risk event possesses a persistent identity and mutable states (e.g., current volume, urgency level).
}

The goal of TingIS is to map an incoming customer incident to either an existing risk event, a newly initialized event, or the null set (noise/suppression).

This mapping is non-trivial due to the ``semantic gap'' between user descriptions and technical root causes. To bridge this gap, we design TingIS based on three core insights. The first is \tb{semantic convergence and identity persistence}, ensuring that incidents originating from the same root cause consistently converge to a unique, persistent ID. The second is a \tb{synergy of hybrid intelligence}, which strategically balances the high cognitive depth of LLMs against the computational cost of processing massive streaming data. This principle of resource awareness is embedded throughout the system: rule-based pre-filtering slashes input volume, LSH and similarity thresholds gate expensive LLM calls, and the use of persistent event states yields asymptotic efficiency gains over time. The third is \tb{multi-constraint SNR balance}, which dynamically suppresses noise by integrating knowledge bases, statistical auditing, and escalation logic.

Guided by these insights, TingIS consists of five orthogonal modules (denoted M1-M5, Figure~\ref{fig:architecture}). Each module is designed to be plug-and-play, allowing for seamless updates - such as integrating more powerful LLMs or faster embedding models - to ensure low-effort maintainability.


\subsection{Semantic Distillation (M1)}

The primary challenge in processing customer incidents is the unstructured, noisy, and colloquially diverse nature of raw user voice. To address this, we implement a \emph{semantic distillation} module to transform raw text into unambiguous semantic units.

Instead of traditional keyword extraction, we leverage an LLM (specifically Qwen3-8B, \citealp{2025Qwen3}) to generate an \emph{initial summary} for every valid incident. This process is governed by a strict prompt constraint: the summary must follow a ``subject + problem'' format (e.g., ``credit card online payment + discount error''), explicitly ignoring emotional expressions, conversational filler, personally identifiable information (PII), and irrelevant details. This strategic design creates a clean, high-density semantic representation at a controlled computational cost. Afterwards, the initial summary is converted into a high-dimensional vector using an embedding model (BGE-M3, \citealp{2024BGE-M3}), serving as the semantic foundation for all downstream operations.

\subsection{Cascaded Routing (M2)}

Production-grade platforms involve numerous business domains that collectively provide exhaustive coverage of all potential customer incidents. Each domain is mapped to a specialized emergency response team accountable for mitigation, uniquely identified by a \emph{business code} (\emph{biz\_code}, an example given in Appendix~\ref{appendix:case-study}). Given the significant semantic divergence across these domains, precise business attribution via routing is a prerequisite for effective discovery. TingIS employs a two-stage routing strategy:

\vspace{0.2em}\noindent\textbf{Keyword-based stage for high-precision:} The system first performs matching against a keyword knowledge base using an ``entity-priority'' principle. If a match is found within the entity fields of the initial summary, the corresponding \emph{biz\_code} is returned immediately. This stage efficiently handles large volumes of clear, well-defined incidents.

\vspace{0.2em}\noindent\textbf{Semantic-based stage for high-recall:} For incidents missing keyword hits, the system performs parallel vector retrieval across multiple vector knowledge bases. Candidates are then refined by a reranker (BGE-Reranker-V2-M3, \citealp{2024BGE-M3}) and filtered via a predefined threshold. \zzy{Candidates accepted by the reranker are routed to the corresponding business domain, while those receiving a low confidence score are dispatched to a fallback domain, where a global control team manually dispatch the incidents.} \yh{Cross-encoder based rerankers achieve superior accuracy via full self-attention but are computationally heavy and cannot pre-compute embeddings~\cite{liao-etal-2024-d2llm}. We meet strict streaming latency constraints by restricting the reranker to a Top-10 vector-retrieved pool.}

\subsection{Event Linking Engine (M3)}

The core challenge in TingIS lies in determining ``event identity'': accurately judging whether multiple incidents, arriving at different times and expressed differently, point to the same underlying risk event. To achieve this, we utilize a \emph{Multi-stage event linking Engine} that follows a progressive refinement process. A detailed illustration of this module is provided in Appendix~\ref{appendix:modules}.

\subsubsection{In-batch Efficient Aggregation}
The system first applies domain constraints by partitioning incidents based on the \emph{biz\_code} provided by M2. Within each partition, we use LSH for high-speed preliminary clustering. To ensure cluster purity, an LLM (Kimi-K2, \citealp{2025Kimi-K2}) performs a representative check on each cluster. \zzy{If a cluster is judged to be impure, the LLM splits it into multiple clusters and generates a title for each one. This synergy of LSH and LLM ensures that the output cluster titles are both comprehensive and mutually exclusive (see Appendix~\ref{appendix:case-study} for an example).}

\subsubsection{Cross-batch Historical Association}
To link current incidents with ongoing events, \zzy{each batch cluster title is embedded and used for retrieval from a historical risk event knowledge base.} We introduce a \emph{time-decay weighting} mechanism to combine semantic similarity with temporal proximity:
\begin{equation}
    s^* = s\cdot e^{-k\Delta t},
\end{equation}
where $s$ is \zzy{the semantic similarity score between the current title embedding and the historical event embedding}, $\Delta t$ is the time (measured in days) since the historical event's last active time, and $s^*$ is the final score. This prevents ``historical inertia,'' where old events might incorrectly absorb new, unrelated incidents. If the highest combined score exceeds a threshold, an LLM performs the final adjudication (merge vs. create new) with a natural language justification. Otherwise, a new risk event is created directly.

\subsection{Event State Management (M4)}
To support real-time risk monitoring and decision-making, we design a layered data model to manage event states and decouple volatility, traceability, and statistical analysis:

\vspace{0.2em}\noindent\textbf{State Layer (Risk Event):} Stores the minimal set of mutable states (e.g., current volume, last altered timestamp, last active timestamp) required for real-time alerting and time-decay calculations.


\vspace{0.2em}\noindent\textbf{Audit Layer (Alert Record):} An immutable log that records the end-to-end evidence chain for every incident (Raw Text $\rightarrow$ Summary $\rightarrow$ Cluster $\rightarrow$ Event ID) and captures every alert trigger, including the context (static thresholds vs. dynamic baselines) and the specific reason for the alert, ensuring 100\% auditability for mis-merges or false alerts and enabling post-mortem analysis of noise reduction strategies.

\vspace{0.2em}\noindent\textbf{Snapshot Layer (Volume Timeline):} Periodically records event volume stock and flow, providing stable, low-cost historical samples for the dynamic baseline calculations in M5 without rescaning heavy logs.

\subsection{Multi-dimensional Denoising (M5)}
Relying solely on volume thresholds often leads to ``alert storms'' during non-failure scenarios (e.g., marketing inquiries). To mitigate this, TingIS integrates three layers of denoising:

\vspace{0.2em}\noindent\textbf{Source Suppression:} During the clustering phase, the system matches clusters against a false-positive sample knowledge base (false-positive KB). If a new cluster is highly similar to historical false positives, it is suppressed before an event is generated.

\vspace{0.2em}\noindent\textbf{Statistical Filtering via Dynamic Baselines:} Incidents must pass a dual-threshold trigger. Beyond static business-level thresholds, an incident's volume must significantly deviate from its \emph{dynamic baseline} ($\mu + 2\sigma$), calculated from the M4 snapshot layer. This filters out periodic business fluctuations.

\vspace{0.2em}\noindent\textbf{Behavioral Constraints:} To prevent alert fatigue, TingIS implements \emph{alert silencing periods}. Once an event is marked as ``In Progress'', further alerts are automatically paused for two hours. However, the system concurrently monitors the slope of the event volume in real-time. If the current volume exhibits an explosive, non-linear surge, the system will bypass the silencing window to implement \tb{alert penetration}, ensuring that critical escalations are immediately delivered to responders despite the ongoing state. A detailed illustration of this module is provided in Appendix~\ref{appendix:modules}.
\section{Experiments}

To comprehensively evaluate TingIS, we establish a layered evaluation framework validating the system through both continuous real-world performance and reproducible offline experiments. Our evaluation is rooted in production data, branching into two complementary paths: (1) \tb{online production validation}, measuring core business impact (Recall and Latency) over a one-month deployment, covering high-priority risk events\footnote{High-priority events refer to those that require immediate attention from SRE (Site Reliability Engineer) teams.} confirmed by expert teams of developers and site reliability engineers (SRE); and (2) \tb{offline benchmark evaluation}, enabling fair, controlled, and reproducible comparisons against baselines and ablation studies.

\begin{table}[th]
    \centering
    \small
    \adjustbox{width=1\linewidth,center}{
    \begin{tabular}{llr}
    \toprule
        \textbf{Dataset} & \textbf{Objective} & \textbf{Size} \\
    \midrule
        \rowcolor{gray!15} \multicolumn{3}{c}{\textit{System-level}}\\
        Alarm Replay Set & End-to-end system behavior simulation & $\sim$50,000 \\
        Benchmark Events & Proxy ground-truth for event discovery & 12 events \\ 
    \midrule
        \rowcolor{gray!15} \multicolumn{3}{c}{\textit{Module-level}}\\
        Event Identity Set & M3 clustering quality & $\sim$1,400 \\
        Routing Set & M2 routing accuracy and coverage & $\sim$3,200 \\
    \bottomrule
    \end{tabular}
    }
    \caption{Overview of Evaluation Datasets.}
    \label{tab:datasets}
\end{table}

\subsection{Datasets and Metrics}

\begin{figure}[t]
    \centering
    \includegraphics[width=1\linewidth]{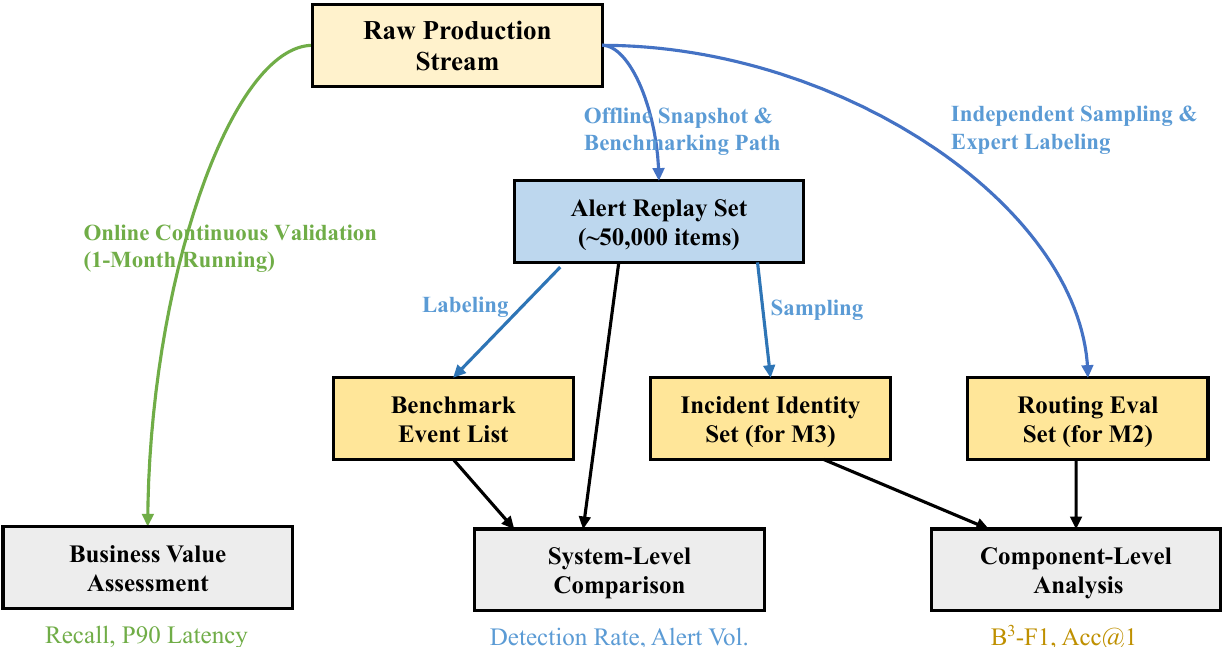}
    \caption{Dataset construction and evaluation metrics.}
    \label{fig:datasets}
\end{figure}

As summarized in Table~\ref{tab:datasets}, we constructed a series of datasets from production snapshots. The \tb{alarm replay set} serves as the parent set to simulate real-world load. From this, we derived the \tb{benchmark events} \zzy{(annotated from 50 thousand incidents by SRE experts)} and the \tb{event identity set} for fine-grained clustering analysis. The \tb{routing set} was independently constructed using a 20\%/80\% split to simulate a ``cold-start'' scenario for evaluating the M2 module's generalization capability. The relation between these datasets are illustrated in Figure~\ref{fig:datasets}. Production performance are measured by risk event recall and 90 percent latency (P90 latency), \zzy{where latency is defined as $t_\text{alert}-t_\text{first\_incident}$}. For system-level benchmarks, we measure risk detection rate and alert volume. For module-level evaluation, we report B$^3$-F1 score and mismerge/fragmentation rates (see Appendix~\ref{appendix:b3_metrics} for more details).

\subsection{Baseline Methods}

We compare TingIS against two groups of methods. Unless otherwise specified, all methods utilize the same M1 \emph{initial summary} as input and share identical embedding and reranking models.

\vspace{0.2em}\noindent\textbf{System-level Baselines} include \emph{keyword-only} (rule-based), \emph{semantic-only} (vector retrieval), \emph{single-stage vector matching} (naive event merging without progressive refinement), and \emph{TingIS w/o Denoising} (same as TingIS but with static M5 thresholds).

\vspace{0.2em}\noindent\textbf{Algorithm-level Baseline} include \emph{generic clustering (DBSCAN)} as a baseline for M3. \yh{To ensure evaluative fairness, the DBSCAN hyperparameters are rigorously optimized via grid search on a held-out validation set.}

\subsection{Results and Analysis}

\subsubsection{System-level Performance and SNR}

TingIS effectively resolves the core conflict between high discovery rates and background noise. In the one-month production run, TingIS achieved a \tb{95\% high-priority incident discovery rate} with a \tb{P90 alert latency of 3.5 minutes}. 

\begin{figure}
    \centering
    \includegraphics[width=1\linewidth]{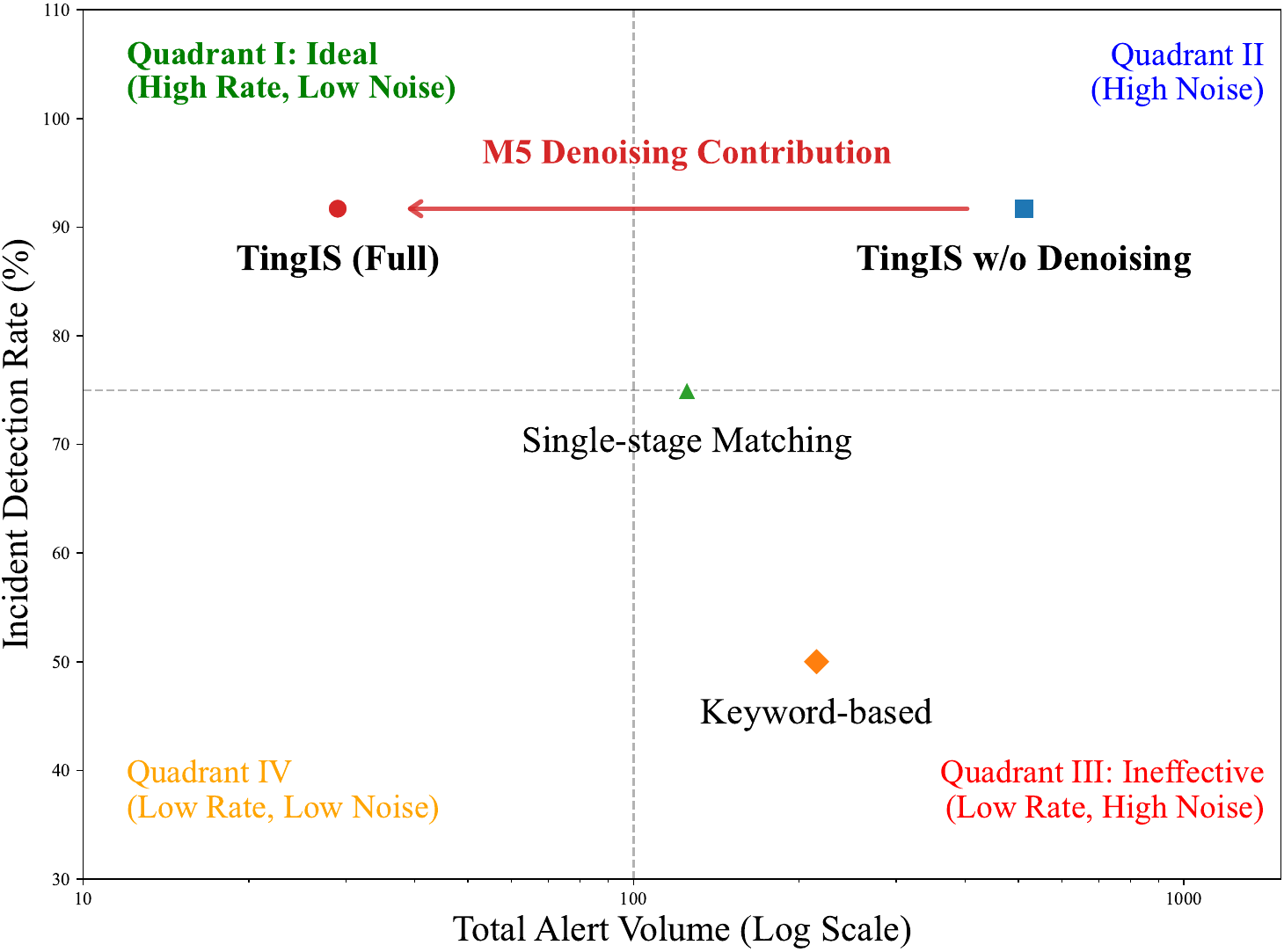}
    \caption{Performance (detection rate vs. alert volume) comparison between different systems.}
    \label{fig:performance}
\end{figure}

Offline results on the alarm replay set (Table~\ref{tab:system_comparison} and Figure~\ref{fig:performance}) show that TingIS significantly reduces noise. While the version without denoising triggered 512 alerts, TingIS suppressed this to 29, representing a \tb{94.3\% noise reduction} with no drop in detection rate. Moreover, its \emph{event-to-alert ratio} of 1.23 (closest to ideal 1.0) confirms the effectiveness of its alert silencing and penetration strategies.

\begin{table}[ht]
    \centering
    \small
    \begin{tabular}{lm{1.3cm}<{\raggedleft}m{2.1cm}<{\raggedleft}}
    \toprule
        \textbf{Method} & \textbf{Total Alerts} & \textbf{Event-to-Alert Ratio} \\
    \midrule
        Keyword-based & 215 & 1.85 \\
        Single-stage Matching & 125 & 1.52 \\
        TingIS w/o Denoising & 512 & 2.18 \\
        \tb{TingIS} & \tb{29} & \tb{1.23} \\
    \bottomrule
    \end{tabular}
    \caption{End-to-End system behavior comparison.}
    \label{tab:system_comparison}
\end{table}

\subsubsection{Event Linking Quality (M1 \& M3)}

The foundation module in TingIS is the event linking Engine (M3). As shown in Table~\ref{tab:clustering}, TingIS leads in B$^3$-F1 (0.826) by achieving a superior balance between ``convergence'' (low fragmentation: 5.8\%) and ``purity'' (low mismerge: 21.5\%). \yh{Operationally, this distinction is critical: a mismerge groups unrelated failures together, leading to fundamentally flawed root-cause analysis and misdirected engineering efforts, whereas slight fragmentation merely creates manageable duplicate workflows. By reducing the disastrous 64.3\% mismerge rate of DBSCAN to an operationally safe 21.5\%, TingIS demonstrates exceptional industrial utility.}


\begin{table}[t]
    \centering
    \small
    \adjustbox{width=\linewidth,center}{
    \begin{tabular}{lrrr}
    \toprule
        \textbf{Method} & \textbf{B$^3$-F1 ($\uparrow$)} & \textbf{Mismerge\% ($\downarrow$)} & \textbf{Frag.\% ($\downarrow$)} \\
    \midrule
        Keyword Grouping & 0.745 & 24.4 & 16.1 \\
        DBSCAN & 0.673 & 64.3 & 5.0 \\
        Vector Matching & 0.744 & 46.3 & 12.0 \\
        \tb{TingIS (Full)} & \tb{0.826} & \tb{21.5} & \tb{5.8} \\
    \midrule
    \end{tabular}
    }
    \caption{Event Identity Quality Comparison, measured by B$^3$-F1 score, mismerge rate, and fragmentation rate.}
    \label{tab:clustering}
\end{table}

\vspace{0.2em}\noindent\textbf{Ablation Study:} Table~\ref{tab:m3_ablation} reveals that \emph{business partitioning} is the cornerstone, as its removal causes a 15.6\% drop in B$^3$-F1. The integration of LLM summary in M1 contributes 7.0\% in B$^3$-F1 and reduces mismerge rate by 66.5\%, while The two-stage LLM application in M3 (intra-batch and final adjudication) also contributes a combined $\sim$5\% B$^3$-F1 improvement and 60\% mismerge reduction. \yh{This empirically validates a profound operational lesson: purely semantic clustering is fundamentally prone to failure in enterprise settings where distinct business domains share similar colloquial vocabularies. Injecting deterministic business metadata acts as an essential, non-negotiable firewall against catastrophic semantic collapse.}

\begin{table}[t]
    \centering
    \small
    \adjustbox{width=\linewidth,center}{
    \begin{tabular}{lll}
    \toprule
        \textbf{Variant} & \textbf{B$^3$-F1} & \textbf{Mismerge \%} \\
    \midrule
        TingIS (Full) & 0.826 & 21.5 \\
        w/o Initial Summary (M1) & 0.768 ($\downarrow$ 7.0\%) & 35.8 ($\uparrow$ 66.5\%) \\
        w/o Business Partition (M3) & 0.697 ($\downarrow$ 15.6\%) & 55.2 ($\uparrow$ 157\%) \\
        w/o Intra-Batch LLM (M3) & 0.796 ($\downarrow$ 3.6\%) & 32.1 ($\uparrow$ 49.3\%) \\
        w/o Final Adjudication (M3) & 0.815 ($\downarrow$ 1.3\%) & 23.9 ($\uparrow$ 11.2\%) \\
    \bottomrule
    \end{tabular}
    }
    \caption{M1 and M3 Ablation Studies.}
    \label{tab:m3_ablation}
\end{table}

\subsubsection{Intelligent Distribution Strategy (M2)}

Analysis of the M2 module yields two key insights (Table~\ref{tab:m2_eval}): (1) \tb{Architecture Over Technique}: The \emph{cascaded} architecture (Acc@1: 0.669) significantly outperforms the \emph{parallel fusion} architecture (Acc@1: 0.460). This confirms that a waterfall strategy prevents noisy keyword results from contaminating the reranker's candidate pool. (2) \tb{Reranker as Risk Controller}: While removing the reranker increases raw Acc@1 to 0.705, it results in 100\% coverage. TingIS (Full) maintains a coverage of 88.1\%, indicating that the reranker acts as a ``quality gatekeeper'' by actively rejecting low-confidence predictions, thereby providing higher SNR input for downstream aggregation.

\begin{table}[t]
    \centering
    \small
    \adjustbox{width=\linewidth,center}{
    \begin{tabular}{lrrr}
    \toprule
        \textbf{Method} & \textbf{Acc@1} & \textbf{Coverage} & \textbf{Latency (s)} \\
    \midrule
        \tb{TingIS (Cascade)} & 0.669 & 0.881 & 53.7 \\
        TingIS (Fusion) & 0.460 & 0.680 & 220.2 \\
        w/o Multi-path Recall & 0.657 & 0.868 & 112.7 \\
        w/o Reranker & 0.705 & 1.000 & 99.2 \\
        Semantic-only & 0.542 & 0.772 & 92.3 \\
        Keyword-only & 0.430 & 0.516 & 4.2 \\
    \bottomrule
    \end{tabular}
    }
    \caption{Intelligent Distribution (M2) Performance.}
    \label{tab:m2_eval}
\end{table}

\zzy{We note that in Table~\ref{tab:m2_eval}, removing the reranker leads to higher accuracy because the knowledge base is comprehensive, containing historical events for all the evaluated incidents. To further verify the role of the reranker, we simulate online scenarios with an incomplete knowledge base by partially removing the database entries. The results (Table~\ref{tab:m2_db}) show that the reranker's value increases as the database degrades. At 60\% database, it improves accuracy while controlling coverage, validating its critical role.}

\begin{table}[t]
    \centering
    \small
    \begin{tabular}{cccc}
    \toprule
        \textbf{DB Size} & \textbf{Reranker} & \textbf{Acc} & \textbf{Coverage} \\
    \midrule
        \multirow{2}{*}{100\%} & w/ & 0.669 & 0.881 \\
         & w/o & 0.705 & 1.000 \\
        \multirow{2}{*}{80\%} & w/ & 0.598 & 0.786 \\
         & w/o & 0.612 & 1.000 \\
        \multirow{2}{*}{60\%} & w/ & 0.534 & 0.686 \\
         & w/o & 0.502 & 1.000 \\
    \bottomrule
    \end{tabular}
    \caption{\zzy{M2 performance at different database (DB) sizes.}}
    \label{tab:m2_db}
\end{table}

\subsection{System Efficiency and Parallelization Analysis}

To meet the high-throughput requirements of enterprise-level production, TingIS implements a deeply parallelized architecture across its pipeline.

\vspace{0.2em}\noindent\textbf{Parallelization Strategy:}~We utilize \texttt{ThreadPoolExecutor}\footnote{\url{https://docs.python.org/3/library/concurrent.futures.html}} to handle concurrent LLM calls and vector searches, including semantic distillation in M1 and cluster auditing in M3. For database operations in M4, we employ batch insertions (\texttt{executemany}) and \texttt{UPDATE CASE} statements to minimize network round-trips and avoid the $N+1$ SQL query problem.

\begin{figure}[t]
    \centering
    \includegraphics[width=1\linewidth]{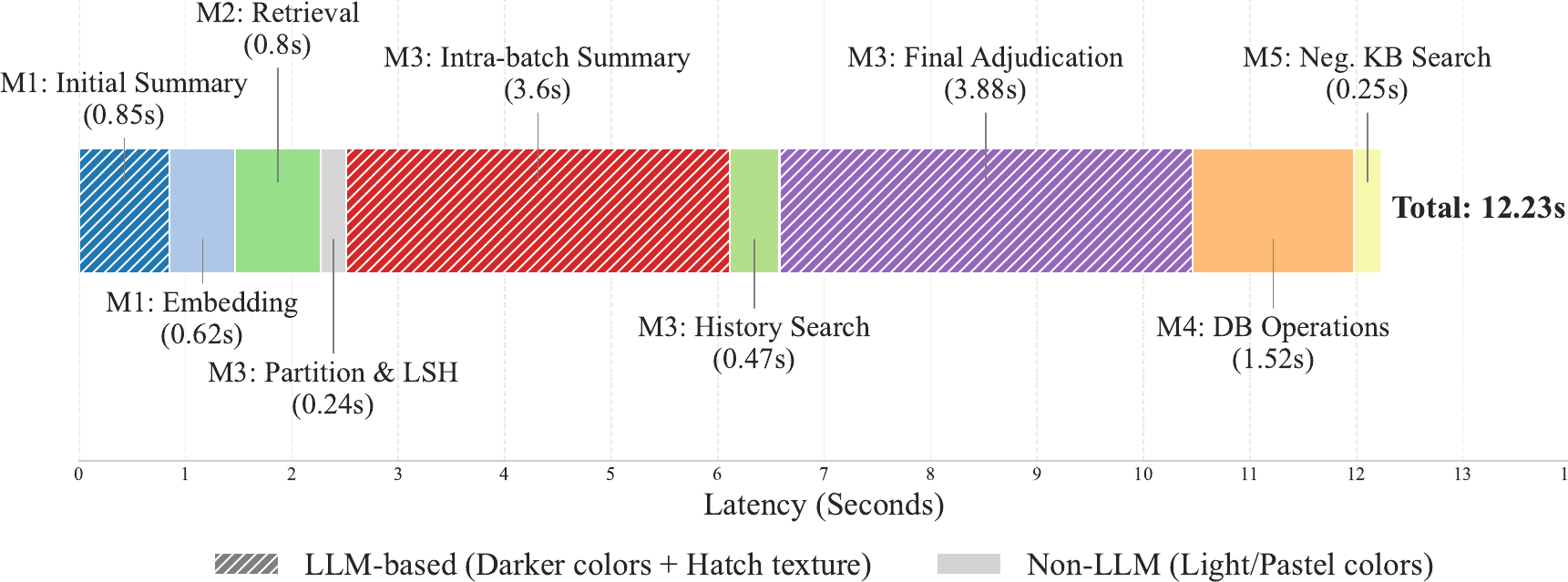}
    \caption{End-to-end latency breakdown.}
    \label{fig:latency}
\end{figure}

\vspace{0.2em}\noindent\textbf{Latency Breakdown:} Our analysis shows an average end-to-end system processing latency of approximately 12.4 seconds per batch. As illustrated by our profiling in Figure~\ref{fig:latency}, LLM-based reasoning (Initial Summary, Intra-batch Summary, and Final Adjudication) remains the primary computational bottleneck, accounting for 8.53 seconds (69.7\% of the total latency). Conversely, non-LLM components, including database operations (1.52s) and vector/keyword retrieval (0.62+0.8+0.47+0.25=2.1s), are highly efficient. This design ensures that even during traffic spikes, the system maintains near real-time processing capabilities with a stable throughput of 2000 queries per minute, while keeping the P90 alert latency below 5 minutes.
In Appendix~\ref{appendix:cost}, we further quantify the computational footprint (8.0M tokens/day) and illustrate how architectural optimizations (e.g., LSH pre-clustering, threshold gating) contain costs within industrial feasibility bounds.

\section{Conclusion}

We present TingIS, an end-to-end risk intelligence system for enterprise-grade incident discovery. TinsIS synergizes LLMs with efficient indexing and historical event association, addressing the challenges of high noise and business heterogeneity inherent in customer incident data. Complemented by cascaded routing, event state management, and multi-dimensional denoising, the system enables stable extraction of actionable risk events from colloquial customer incidents. 

Deployed in a large-scale fintech environment, TingIS processes over 300,000 incidents daily and 2,000 per minute with a 95\% discovery rate for high-priority incidents and a P90 alert latency of 3.5 minutes. Benchmark results also demonstrate that our hybrid intelligence approach significantly improves SNR and reduces false alerts, event mismerge, and event fragmentation.

Beyond technical contributions, TingIS embodies hard-won operational insights from real-world deployment. In Appendix~\ref{appendix:lessons}, we further provide actionable guidance for industrial NLP practitioners facing similar constraints by documenting critical lessons on handling data skew, designing robust routing strategies, and integrating LLMs responsibly.


\bibliography{custom}

\appendix

\clearpage
\section{Case Studies}
\label{appendix:case-study}

To demonstrate the real-world efficacy of TingIS, we analyze three representative cases from our production environment in Table~\ref{tab:case-study}, focusing on efficiency, semantic convergence, and denoising. \zzy{In Table~\ref{tab:biz_code}, we also provide an anonymized example of the \emph{biz\_code} taxonomy.}

\section{Related Work}

This section reviews prior work related to TingIS from four perspectives: event detection from text streams, LLM-based text clustering, multi-dimensional denoising for customer incidents, and domain-adaptive routing. We emphasize the limitations of existing approaches in real-time, noisy, and heterogeneous enterprise settings, and clarify how TingIS advances the state of the art.

\subsection{Streaming Event Detection from Text}

Early streaming clustering methods such as CluStream~\citep{2003clustream} and DenStream~\citep{2006denstream} focus primarily on numerical features and density evolution, making them ill-suited for semantically rich text streams.  
With the rise of social media, researchers began incorporating text embeddings into streaming event detection. \citet{li2019realtime} introduces sliding-window-based clustering for Twitter streams, enabling near-real-time detection of emerging topics. However, the use of fixed windows often fragments long-lived events across temporal boundaries.  To mitigate this issue, Embed2Detect~\citep{2022embed2detect} utilizes word embeddings for event detection in social media streams, enabling semantics-aware detection of temporal clusters, while \citet{2021entity-aware} employs entity-aware contextual embeddings for online news stream clustering.

However, all these methods treat events as transient clusters and do not explicitly model long-term event identity, especially under semantic drift and evolving customer incidents. In contrast, TingIS combines time-decayed similarity with LLM-based adjudication to explicitly achieve event identity persistence.

\subsection{Large Language Models for Text Clustering}

Recent studies in text clustering have increasingly explored the integration of LLMs to enhance semantic representation and clustering performance. \citet{2024few-shot-clustering} improves semi-supervised text clustering by incorporating LLM guidance at multiple stages of the clustering pipeline, including feature enrichment, constraint generation, and post-clustering correction. Complementing this, \citet{2025clustering-with-llm-embedding} empirically shows that LLM embeddings capture richer linguistic nuances compared to traditional representations, leading to better cluster purity. Alternative frameworks have recast clustering as a classification problem using in-context learning, bypassing the need for conventional clustering algorithms while achieving competitive performance~\citep{2025clustering-as-classification}. Other research has proposed Context-Aware Clustering with LLMs that leverage attention and supervised losses to scale clustering to large entity sets effectively~\citep{2024context-aware-clustering}, and recent work on in-context clustering highlights LLMs’ zero-shot capabilities for capturing complex relationships in data~\citep{2025in-context-clustering}. Together, these studies illustrate a rapidly evolving landscape where LLMs not only provide semantically rich embeddings for traditional clustering algorithms but also enable novel paradigms for clustering via prompting, and few-shot learning. However, using LLMs alone leads to high computational costs and latency, whereas TingIS dynamically combines LLM-based clustering with more efficient components such as Locality-Sensitive Hashing.

\begin{table}[t]
    \centering
    \small
    \adjustbox{width=1\linewidth,center}{
    \begin{tabular}{lll}
    \toprule
        \textbf{Level} & \textbf{Description} & \textbf{Examples} \\
    \midrule
        L1 & Business Group & Digital Finance, Digital Payments \\
        L2 & Product Line & Insurance, Consumer Credit \\
        L3 & Sub-product & Health Insurance, Mutual Funds \\
        L4 & Feature/Scenario & Claims Processing, Fund Redemption \\
    \bottomrule
    \end{tabular}
    }
    \caption{\zzy{An anonymized example of the four-level \emph{biz\_code} taxonomy.}}
    \label{tab:biz_code}
\end{table}

\begin{table*}[t]
    \centering
    \small
    \adjustbox{width=1\textwidth,center}{
    \begin{tabular}{m{5.5cm}m{6cm}m{6cm}}
    \toprule
        \tb{Event Description} & \tb{System Behavior} & \tb{Analysis} \\
    \midrule
        \rowcolor{gray!15} \multicolumn{3}{c}{\textit{Case A: Rapid Capture of Instant Risks (Efficiency)}}\\
        During a peak transaction window of a major promotional event, a core payment gateway experienced transient instability. & Utilizing the parallelized architecture of Layer II, TingIS completed the cycle from the first customer incident to the final alert trigger within 2 minutes. & Compared to traditional periodic batch-processing (5–15 min), TingIS provided a critical ``golden window'' for SRE teams to mitigate the fault before escalation. \\
    \midrule
        \rowcolor{gray!15} \multicolumn{3}{c}{\textit{Case B: Convergence of Diverse Expressions (Semantic Alignment)}}\\
        Following a version update of a virtual pet feature, users reported: ``my pet won't sleep'', ``the sleep button is unresponsive'', and ``the game is stuck on the loading screen''. & M1 normalized these into a core summary (``Virtual Pet + Function Failure''). M3 linked these disparate reports to a single historical risk event ID via cognitive adjudication. & A keyword-based system would likely fragment these into low-volume ``minor issues,'' failing to trigger a high-priority alert. \\
    \midrule
        \rowcolor{gray!15} \multicolumn{3}{c}{\textit{Case C: Suppression of High-Volume Inquiries (Denoising)}}\\
        During a monthly social campaign, inquiries regarding ``how to check reward progress'' surged to 20x the baseline volume within 10 minutes. & M5 identified the cluster as highly similar to a ``historical inquiry'' entry in the False-Positive KB. M4's dynamic baseline confirmed this surge matched expected social campaign patterns. & By distinguishing high-volume non-risk inquiries from actual failures, TingIS effectively prevents ``alert fatigue'' for the emergency response teams. \\
    \bottomrule
    \end{tabular}
    }
    \caption{Case studies of TingIS's behavior in production environments.}
    \label{tab:case-study}
\end{table*}

\subsection{Multi-dimensional Denoising of Customer Incidents}

Noise reduction is a long-standing challenge in customer service analytics. Rule-based filtering and knowledge-guided denoising rely on manually curated keyword lists or pattern libraries~\citep{2021keyword-filtering}, which are brittle under domain drift and emerging issues. Statistical anomaly detection methods focus on identifying volume deviations in time series data~\citep{2009statistic-anomoly-1,2009statistic-anomoly-2}, while \citet{2019anomoly-detection-microsoft} employs spectral residual and convolutional neural networks to improve the performance. These methods are effective for monitoring numerical data, but lack semantic awareness and cannot distinguish between genuine failures and high-volume non-risk inquiries.
In comparison, TingIS introduces a complaint-specific denoising funnel that integrates semantic false-positive matching, dynamic baselines, and behavioral constraints such as alert silencing with slope penetration, achieving high noise reduction without sacrificing recall for high-priority incidents.

\subsection{Domain-adaptive Routing and Cascaded Retrieval}

Hybrid retrieval architectures that combine sparse and dense representations have proven effective in improving recall and robustness~\citep{2020DPR}. \citet{2017cascade-ranking} operationalizes this practice in E-commerce search, introducing a cascade ranking model to balance accuracy and latency, while \citet{2025hybrid-ai-conversation} applies such hybrid systems in customer support applications. However, existing systems often ignore domain heterogeneity, leading to cross-domain noise propagation and cold-start failures.

TingIS extends cascaded retrieval by explicitly incorporating business-domain isolation. Its waterfall routing strategy—keyword matching, multi-path vector recall, and reranker-based quality control—ensures high precision in head cases and robust coverage in long-tail scenarios, even under cold-start conditions.

\section{Detailed Illustration of System Modules}\label{appendix:modules}

\begin{figure}[th]
    \centering
    \includegraphics[width=1\linewidth]{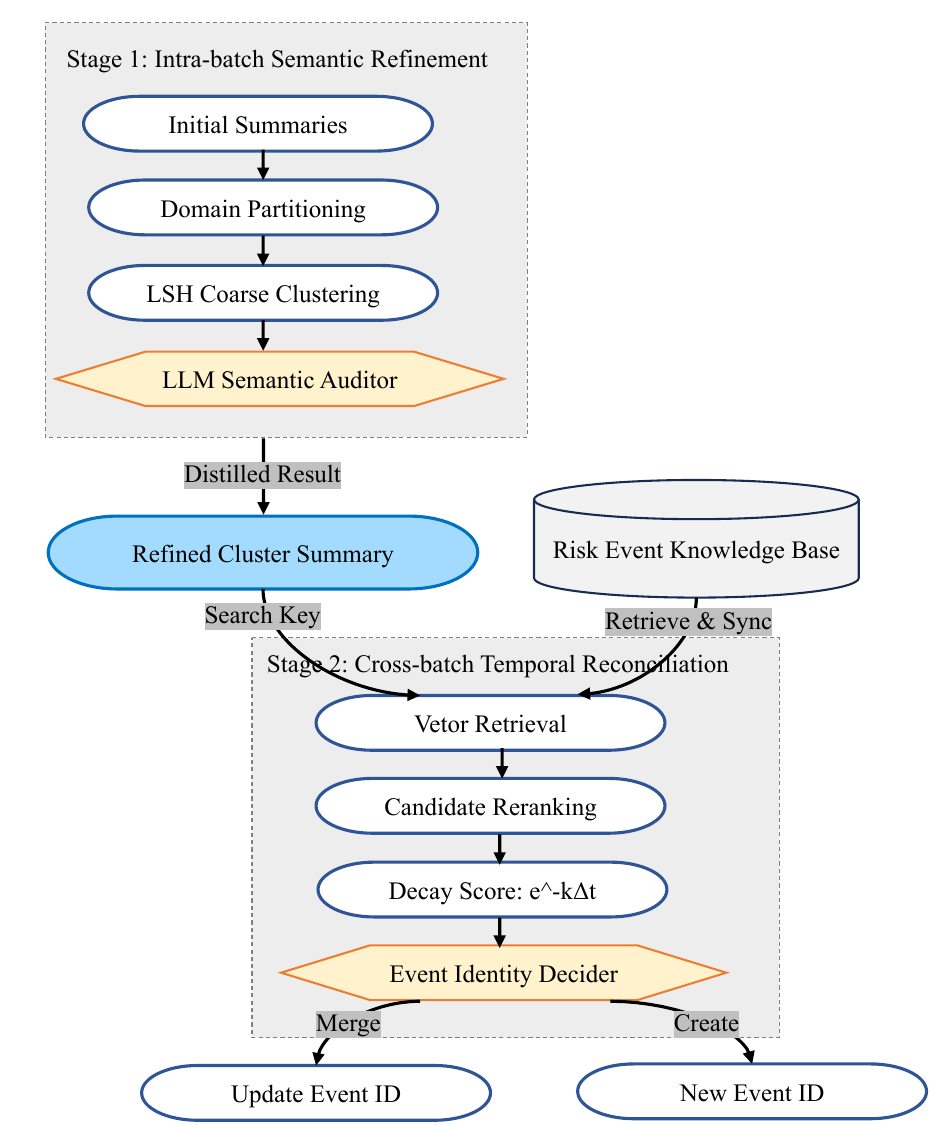}
    \caption{Multi-stage event linking engine (M3) in TingIS.}
    \label{fig:m3}
\end{figure}

\begin{figure*}[th]
    \centering
    \includegraphics[width=1\textwidth]{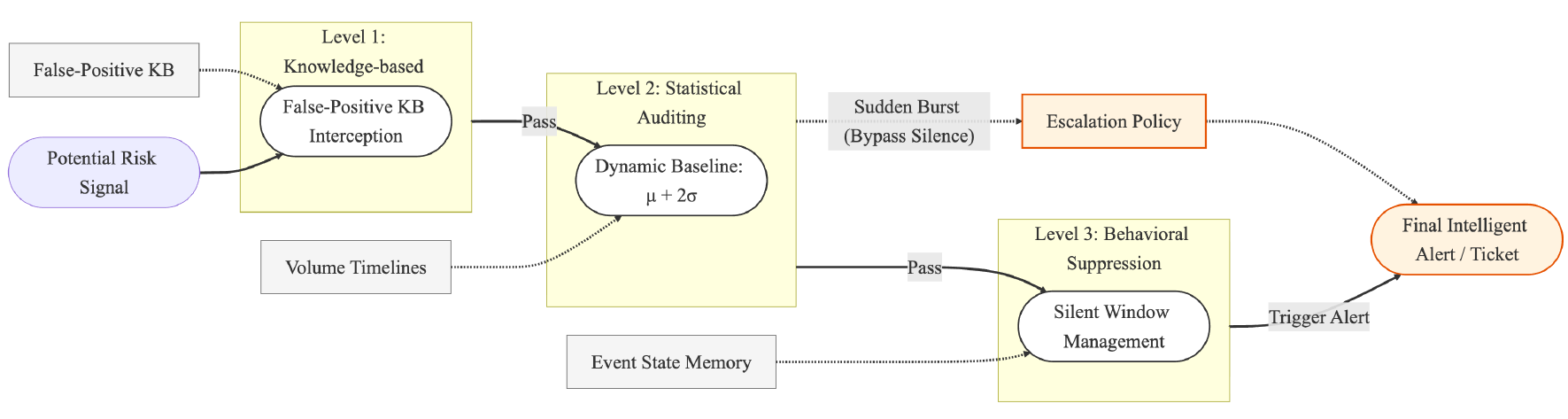}
    \caption{Multi-dimensional denoising module (M5) in TingIS.}
    \label{fig:m5}
\end{figure*}

We provide more detailed illustrations of the event linking engine (M3) and denoising pipeline (M5) in Figure~\ref{fig:m3}, \ref{fig:m5}, respectively.

\section{B$^3$ Evaluation Metrics}\label{appendix:b3_metrics}

To evaluate the quality of the event linking Engine (M3), we employ the B-Cubed (B$^3$) precision, recall, and F1-score. Unlike standard classification scores that rely on predefined category labels, B$^3$ metrics are designed for clustering tasks where cluster IDs are arbitrary and do not directly map to ground-truth labels.

\subsection{Mathematical Definition}
Let $N$ be the total number of incidents in the evaluation set. For any incident $i$, let $L(i)$ be the set of incidents that share the same ground-truth event label as $i$, and $C(i)$ be the set of incidents assigned to the same cluster as $i$ by the system.

The B$^3$ precision ($P$) and recall ($R$) are defined by calculating the per-item precision and recall and then averaging them across all items:

\begin{equation}
P(i) = \frac{|C(i) \cap L(i)|}{|C(i)|}
\end{equation}

\begin{equation}
R(i) = \frac{|C(i) \cap L(i)|}{|L(i)|}
\end{equation}

The overall system performance is the mean of these individual scores:

\begin{equation}
\text{B}^3\text{-Precision} = \frac{1}{N} \sum_{i=1}^{N} P(i)
\end{equation}

\begin{equation}
\text{B}^3\text{-Recall} = \frac{1}{N} \sum_{i=1}^{N} R(i)
\end{equation}

The B$^3$-F1 score is the harmonic mean of the aggregate Precision and Recall:
\begin{equation}
\text{B}^3\text{-F1} = \frac{2 \cdot \text{B}^3 \text{ Precision} \cdot \text{B}^3 \text{ Recall}}{\text{B}^3 \text{ Precision} + \text{B}^3 \text{ Recall}}
\end{equation}

\subsection{Interpretation in Incident Discovery}
In the context of TingIS, these metrics provide fine-grained insights into the clustering behavior:
\begin{itemize}
    \item \textbf{Precision (Purity):} A high B$^3$ precision indicates a low \textit{Mismerge Rate}. It means that most incidents within the same generated cluster actually belong to the same underlying risk event.
    \item \textbf{Recall (Completeness):} A high B$^3$ recall indicates a low \textit{Fragmentation Rate}. It means that incidents belonging to the same root cause are successfully converged into a single persistent event ID rather than being scattered across multiple clusters.
\end{itemize}

By using B$^3$-F1 instead of standard F1, we avoid the ``label matching'' problem, ensuring that the evaluation focuses on the \textit{relationships} between incidents rather than the specific naming of cluster IDs.

\section{Resource and Cost Efficiency Analysis}
\label{appendix:cost}

Building upon the design principles validated in Appendix~\ref{appendix:lessons} (e.g., rule-based pre-filtering, fixed-size batching), we quantify TingIS's computational footprint using one month of production monitoring data (daily median input: 250k customer incidents). All metrics reflect live operational behavior.

\subsection{Input Volume Reduction via Lightweight Preprocessing}
As established in Appendix~\ref{appendix:lessons} (Section 4.1), rule-based filtering eliminates low-signal customer incidents while preserving $>99\%$ recall for high-priority incidents. This reduces downstream processing volume to $\sim$50k customer incidents/day at near-zero computational cost—establishing the foundational efficiency layer.

\subsection{LLM Token Consumption: Quantitative Breakdown}
Token usage is strictly monitored across LLM-dependent stages (Table~\ref{tab:token_consumption}). Crucially, \textit{no LLM is invoked on filtered customer incidents}, and Kimi-K2 calls are minimized via algorithmic gating validated in Appendix~\ref{appendix:lessons} (Sections 4.2, 4.3).

\begin{table*}[th]
\centering
\small
\adjustbox{width=1\textwidth}{
\begin{tabular}{lccc}
\toprule
\textbf{Module} & \textbf{Function} & \textbf{Daily Tokens} & \textbf{Optimization Mechanism} \\
\midrule
M1 (Qwen3-8B) & Semantic distillation & 5.0M & Fixed summary ($\sim$100 tokens: 80 input [prompt+text] + 20 output) \\
M3 (Kimi-K2) & refinement \& adjudication & 3.0M & LSH pre-clustering + $s^* > 0.95$ threshold bypass \\
\midrule
\textbf{Total} & & \textbf{8.0M} & \\
\bottomrule
\end{tabular}
}
\caption{Daily token consumption (median values). Optimization mechanisms directly implement design choices from Appendix~\ref{appendix:lessons}.}
\label{tab:token_consumption}
\end{table*}

\begin{itemize}
    \item \textbf{M1 Efficiency:} Processes all $\sim$50k filtered customer incidents. Strict prompt engineering enforces concise summaries ($\sim$100 tokens total), yielding 5.0M tokens/day.
    \item \textbf{M3 Efficiency:} Kimi-K2 is invoked once per LSH-generated cluster ($\sim$30,\!000 clusters/day: 250 batches $\times$ 10 biz\_codes $\times$ 12 clusters/biz\_code). The $s^* > 0.95$ threshold bypasses LLM adjudication for $>70\%$ of historical matches during steady-state operation, containing total consumption at 3.0M tokens/day.
\end{itemize}

\subsection{End-to-End Cost per Actionable Alert}
After multi-dimensional denoising (M5), TingIS generates $\sim$29 high-confidence alerts/day (validated by SRE teams). The \textit{end-to-end computational cost per actionable alert} is:
\begin{equation}
    \frac{\text{Total Tokens}}{\text{Validated Alerts}} = \frac{8.0\text{M}}{29} \approx 275\text{K} \text{ tokens/alert}.
\end{equation}
This metric holistically captures the full pipeline cost—from raw user voice ingestion to human-actionable alert—including all intermediate processing of non-alerting incidents.

\subsection{Quantified Impact of Design Choices}
Three architecture decisions from Appendix~\ref{appendix:lessons} directly enable sustainable scaling:
\begin{enumerate}
    \item \textbf{Rule-Based Pre-Filtering (Appendix~\ref{appendix:lessons}, Sec 4.1):} Eliminates $\sim$200k low-signal customer incidents/day, reducing downstream LLM load by $80\%$.
    \item \textbf{Fixed-Size Batching (Appendix~\ref{appendix:lessons}, Sec 4.2):} Stabilizes Kimi-K2 invocation rate at $\sim$30k clusters/day (vs. volatile time-window batching), preventing RPM throttling and OOM risks.
    \item \textbf{Threshold-Gated LLM Adjudication (Appendix~\ref{appendix:lessons}, Sec 4.3):} The $s^* > 0.95$ rule bypasses LLM calls for $>70\%$ of historical matches, containing daily Kimi-K2 tokens at 3.0M. Monitoring curves show consumption declines progressively from cold-start peaks to a stable baseline.
\end{enumerate}

\noindent\textbf{Operational Viability.} While exact Kimi-K2 API costs are subject to commercial agreements, the sustained 8.0M tokens/day operational scale is tractable for enterprise deployment. Critically, TingIS \textit{proactively minimizes} token consumption through architectural design—transforming LLMs from a cost liability into a targeted, high-value component. This quantifiable efficiency validates the design philosophy articulated in Appendix~\ref{appendix:lessons}.

\section{Lessons Learned and System Iteration}\label{appendix:lessons}

Through iterative deployment and validation of TingIS in a high-stakes production environment, we distilled the following empirically grounded lessons. Each insight addresses concrete challenges observed during scaling, with methodological rigor suitable for industrial NLP system design.

\subsection{Data Preprocessing: Rule Filtering Requires Recall-Aware Validation}
\textbf{Observation:} Customer incident distribution exhibited severe skew 73\% concentrated across 8 high-frequency business domains; >50\% comprised low-information content such as emotional expressions or generic inquiries). Naive filtering risked discarding actionable signals.\
\textbf{Solution:} Six configurable filtering rules (length thresholds, prefix+length patterns, keyword logic combinations) were designed and rigorously validated against historical fault logs to guarantee zero degradation in high-priority incident recall. Daily customer incident volume reduced from 
$\sim$250k to $\sim$50k (80\% filtered).\
\textbf{Insight:} Lightweight rule-based preprocessing is indispensable for computational efficiency, but rule thresholds must be empirically anchored to business-critical recall metrics—not heuristic assumptions. Validation against historical faults is non-negotiable.

\subsection{Batching Strategy: Fixed-Size Batching Ensures Operational Stability}
\textbf{Observation:} Customer incident streams displayed high temporal variance (peak-to-trough ratio >100×). Time-window batching induced dual failures during traffic surges: (1) local resource exhaustion (OOM risks) and (2) downstream service throttling due to exceeding RPM (requests per minute) quotas of LLM/embedding APIs; conversely, it caused severe resource underutilization during low-traffic periods.\
\textbf{Solution:} Fixed-size batching ({batch size=200}) was adopted. This enforces constant per-batch computational load, providing natural backpressure and stabilizing SLAs for downstream LLM and embedding services.\
\textbf{Insight:} In non-stationary traffic regimes, fixed-size batching is a robust operational choice that prioritizes system resilience over theoretical elegance. Design must explicitly accommodate downstream service constraints.

\subsection{Business Routing: Keywords Demand Cross-Domain Discriminative Design}
\textbf{Observation:} Initial dual-path routing (keyword matching + vector retrieval) required refinement to minimize cross-domain contamination while maintaining coverage.\
\textbf{Solution:} Keywords were engineered with explicit cross-domain discriminative power (e.g., disambiguating terms for account inquiry'' versus transaction failure'' domains). Multi-domain customer incidents retained top-3 business domains; customer incidents exceeding this threshold were filtered (validated by historical analysis: 
>98\% of valid customer incidents exhibit clear domain attribution). Campaign-specific domains were pre-configured; vector indices were incrementally updated using Global Operations Center (GOC)-verified bad cases.\
\textbf{Insight:} Keywords function as explicit semantic anchors requiring cross-domain discriminative design. Routing systems achieve sustained precision through a \textit{lightweight closed-loop feedback mechanism}: GOC-verified anomalies trigger \textit{targeted, incremental updates} to keyword libraries and vector indices, balancing automation with minimal operational overhead. This design ensures knowledge bases evolve efficiently without demanding continuous manual intervention.

\subsection{Clustering Quality: LLMs Enable Critical Semantic Disentanglement}
\textbf{Observation:} Pure embedding-based clustering suffered structural ambiguity—overweighting problem'' tokens (e.g., failure'') while neglecting subject'' distinctions. \textit{Example:} merging marketing campaign reward redemption failure'' and NFC payment functionality reward usage failure'' (root causes reside in campaign logic versus payment pipeline).
\textbf{Solution:} LLM-generated structured summaries (subject + problem'', e.g., ``reward + redemption failure'') disentangled semantics. \
\textbf{Insight:} LLMs are indispensable for bridging the colloquial-to-technical semantic gap in incident description. 

\subsection{Cross-Cutting Principles for Industrial NLP}
\noindent Synthesizing domain-specific lessons, we distill three principles with broad applicability to industrial NLP system design:
\begin{itemize}
    \item \textbf{Validation Over Heuristics:} Rule thresholds and batching strategies must be empirically validated against historical fault logs—not theoretical assumptions—to preserve critical signal integrity.
    \item \textbf{Knowledge Requires Continuous Curation:} Keyword libraries and vector indices decay without closed-loop feedback; operational sustainability demands lightweight, targeted refinement mechanisms.
    \item \textbf{Transparency in Failure Analysis:} Documenting limitations (e.g., pure embedding clustering's subject-blindness) complements success metrics by providing actionable insights into failure modes, strengthening methodological credibility and community trust.
\end{itemize}

\end{document}